\documentclass[conference,a4paper]{APSIPA2022}
\usepackage{amsmath}
\usepackage[psamsfonts]{amssymb}
\usepackage{amsxtra}
\usepackage{hyperref}
\usepackage{cite}
\usepackage{pdfpages}
\usepackage{algorithm}
\usepackage{algpseudocode}
\usepackage{float, subfig}
\usepackage[left=1.3cm,right=1.35cm,top=1.9cm,bottom=4.3cm]{geometry}

\usepackage{fancyhdr}
\fancypagestyle{firstpage}{%
  \lfoot{978-616-590-477-3 $\copyright$  2022 APSIPA}
  \rfoot{APSIPA ASC 2022}
}

\begin{document}

\title{A novel deep learning-based approach for sleep apnea detection using single-lead ECG signals}

\author{
    \authorblockN{Anh-Tu Nguyen\authorrefmark{1}\authorrefmark{3}, 
    Thao Nguyen\authorrefmark{1}\authorrefmark{3}, 
    Huy-Khiem Le\authorrefmark{1},
    Huy-Hieu Pham\authorrefmark{1}\authorrefmark{2}, 
    and Cuong Do\authorrefmark{1}\authorrefmark{2}
    }

    \authorblockA{\authorrefmark{1}
    College of Engineering \& Computer Science, VinUniversity, Hanoi, Vietnam\\}
    
    \authorblockA{\authorrefmark{2}
    VinUni-Illinois Smart Health Center, VinUniversity, Hanoi, Vietnam\\}
    
    \authorblockA{\authorrefmark{3}
    These authors contributed equally to this work.\\}
    }

\maketitle
\thispagestyle{empty}

\begin{abstract}

    Sleep apnea (SA) is a type of sleep disorder characterized by snoring and chronic sleeplessness, which can lead to serious conditions such as high blood pressure, heart failure, and cardiomyopathy (enlargement of the muscle tissue of the heart). The electrocardiogram (ECG) plays a critical role in identifying SA since it might reveal abnormal cardiac activity. Recent research on ECG-based SA detection has focused on feature engineering techniques that extract specific characteristics from multiple-lead ECG signals and use them as classification model inputs. In this study, a novel method of feature extraction which based on the detection of S peaks is proposed to enhance the detection of adjacent SA segments using a single-lead ECG. In particular, ECG features collected from a single lead (V2) are used to identify SA episodes. On the extracted features, a CNN model is trained to detect SA. Experimental results demonstrate that the proposed method detects SA from single-lead ECG data is more accurate than existing state-of-the-art methods, with 91.13\% classification accuracy, 92.58\% sensitivity, and 88.75\% specificity. Moreover, the further usage of features associated with the S peaks enhances the classification accuracy by 0.85\%. Our findings indicate that the proposed machine learning system has the potential to be an effective method for detecting SA episodes.
    
\end{abstract}

\section{Introduction}
     Sleep apnea (SA) is the most prevalent breathing problem associated with sleep~\cite{javaheri2017sleep}. It causes people to repeatedly stop and start breathing during sleep. There are various types of sleep apnea, but obstructive sleep apnea is the most prominent, which occurs when upper airway muscles relax during sleep and obstruct the airway, preventing adequate airflow~\cite{javaheri2017sleep}. Patients' breathing may stop for 10 seconds or longer before the reflexes kick in and then resume breathing. 
     Sleep apnea affects approximately 3\% of normal-weight adults, but over 20\% of obese individuals, and in general, men are more susceptible to sleep apnea than women~\cite{johns-hopkins-medicine-2021}.\\
    
    Sleep apnea is frequently associated with heart disease and metabolic disorders such as diabetes~\cite{leung2001sleep}. Several studies have demonstrated a link between sleep apnea and health issues such as type 2 diabetes, strokes, heart attacks, and even a shorter lifespan~\cite{leung2001sleep, young2009burden, aronsohn2010impact, worsnop1998prevalence}.
    It is essential to diagnose and detect sleep apnea in order to avoid long-term health repercussions. The questionnaire, which includes the STOP-Bang Questionnaire~\cite{chung2016stop, nagappa2015validation} and Berline Questionnaire~\cite{netzer1999using}, is one method for screening patients at risk for SA syndrome. The gold standard for diagnosing sleep-related breathing disorders, however, is polysomnography --- a method for collecting physiologic parameters during sleep~\cite{rundo2019polysomnography}. A polysomnogram (PSG) is a diagnostic test that uses electroencephalogram (EEG), electrooculogram (EOG), electromyogram (EMG), electrocardiogram (ECG), and pulse oximetry, in addition to airflow and respiratory effort, to identify the underlying reasons of sleep abnormalities~\cite{rundo2019polysomnography}. However, this diagnostic procedure is time-consuming, expensive, and inconvenient. Patients must be connected to at least 22 electrodes for several nights in order to measure 11 channels of sleep signals in the laboratory using specialized equipment~\cite{sharma2018application}. This is a barrier for patients to independently install and use the device at home. Moreover, to diagnose a patient, physicians must spend considerable time monitoring and interpreting that data. Consequently, the PSG-based technique is costly, complicated, and cumbersome. That trigger a need for a simple, affordable, and user-friendly alternative. \\
    
    ECG is a viable tool for diagnosing and detecting sleep apnea, which has lately generated considerable interest. While some study groups have demonstrated that a patient's cardiovascular activity changes during sleep apnea, resulting in typical ECG readings~\cite{cloward2003left, narkiewicz1998altered}, other studies have demonstrated a link between patients with SA and other cardiovascular disorders~\cite{shamsuzzaman2003obstructive}. Thus, the Apnea-ECG database~\cite{penzel2000apnea} was developed by PhysioNet to spot abnormalities in patients' ECG signals when they have SA. Various research use ECG data to identify SA episodes, some of which employ traditional machine learning techniques based on feature engineering.~\cite{feng2020sleep, bahrami2022sleep, shen2021multiscale, pombo2020classifier, fatimah2020detection, singh2019novel, pinho2019towards} and others applying deep learning techniques~\cite{dey2018obstructive, bahrami2022sleep} for their classification tasks. In~\cite{li2018method}, Li \textit{et al.} suggested a technique for detecting SA using ECG data, which is based on a deep neural network and a Hidden Markov model (HMM). The approach used a sparse auto-encoder to learn features, which were then fed into two types of classifiers (SVM and ANN). In the per-segment SA detection, they achieve 84.5\% classification accuracy. In~\cite{pinho2019towards}, a Sgolay filter was applied to extract the Heart Rate Variability (HRV) and the ECG-Derived Respiration (EDR), then they were used for the training procedure to achieve 82.2\% accuracy. Shen \textit{et al.} introduced a method that based on a weighted-loss time-dependent (WLTD) classification model and a multiscale dilation attention 1-D convolutional neural network (MSDA-1DCNN)~\cite{shen2021multiscale}. That study obtained 89.4\% accuracy, 89.8\% sensitivity and 89.1\% specificity. A novel method for diagnosing SA using a pre-trained AlexNet model is reported in~\cite{singh2019novel}, in which per-minute segments of a single-lead ECG recording are decomposed using continuous wavelet transform (CWT) and subsequently 2D scalogram images are created. Following that, a CNN based on a deep learning algorithm is used to improve classification performance. \\
    
    While recent studies have sought to utilize ECG signals as direct input to deep learning models, research using handcrafted features offers the potential for development because of its transparency and interpretability. However, recent research targeted at identifying SA by feature engineering only consider the position of the R peak in the ECG, neglecting the remaining four peaks (P, Q, S, and T). In this work, we not only identify the R peak, but also determine the position of the S peak and demonstrate that using the S peak enhances the model's performance. To reduce noise and signal artifacts, we first applied a Finite Impulse Response (FIR) band-pass filter to the signal. Afterward, features are extracted based on the detection of the R and S peaks. We employ the feature extraction approach given by Wang \textit{et al.}~\cite{wang2019sleep}, in which an ECG record is divided into 5 minute-long segments, then the R peak is detected. In this work, the detection of S peak is added, followed by cubic interpolation to generate 900 values for each feature. The extracted features were used as input to the SE-ResNext~50 model to classify ECG signals with and without SA. Our model achieved 91.13\% classification accuracy, 92.58\% sensitivity, and 88.75\% specificity. It has been demonstrated that using additional features extracted from S peaks improves accuracy by 0.85\% compared to using only R-peak-related features, which may indicate anomalies in the S peak morphology during SA. 
    
    The rest of this paper is structured as follows. Methods for ECG signal preprocessing, feature extraction, and classification are introduced in section~\ref{section:methods}. Section~\ref{section:experiments} describes the dataset and experimental setups. In Section~\ref{section:experiments} and ~\ref{section:conclusion} we report experimental results and summarize the key findings of this work. 
    
\section{Methods}
    \label{section:methods}
\begin{figure}[!h]
            \centering
            \includegraphics[scale=0.35]{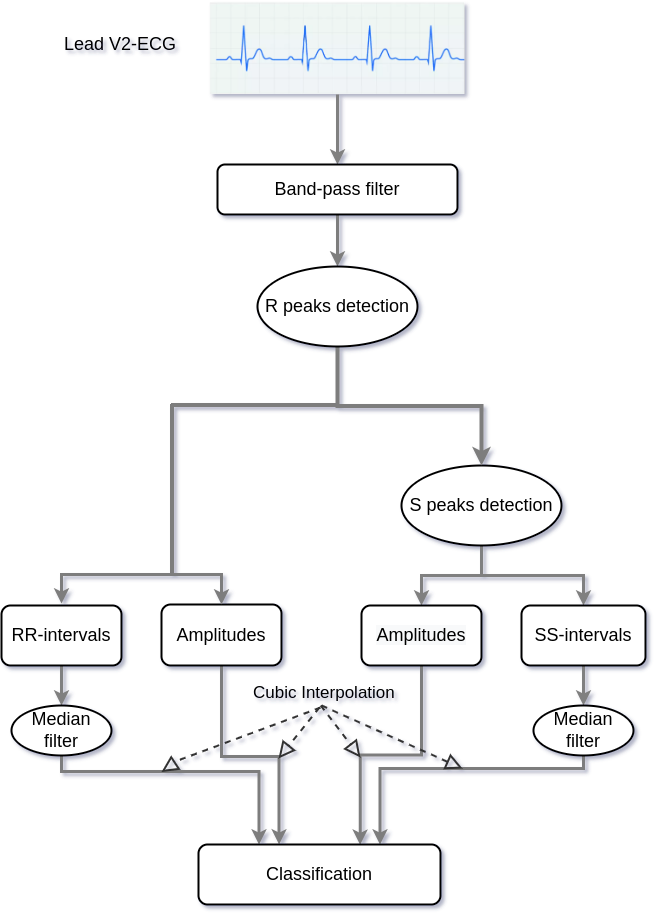}
            \caption{An overview of the proposed approach for detecting sleep apnea.}
            \label{fig:scheme}
    \end{figure}

    Methods for preprocessing the ECG signal, extracting hand-crafted features, and developing a classifier for ECG with and without SA are described in this section.
    The proposed system's schematic is depicted in Fig.~\ref{fig:scheme}. 
    
    \subsection{Pre-processing data}
\begin{figure}[hbt!]
          \centering
          \subfloat[]{\includegraphics[scale=0.6]{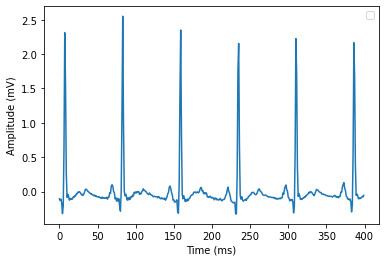} 
          \label{subfig: Subfig3a}} \\
          \subfloat[]{\includegraphics[scale=0.6]{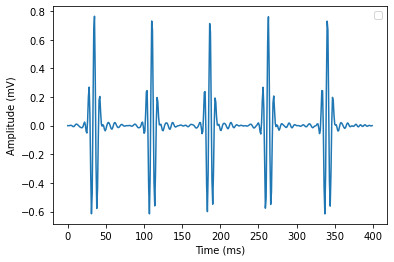}
          \label{subfig: Subfig3b}}
          \caption{\textit{(a) }The original signal of V2 lead ECG and \textit{(b)} The component falls within the 8 to 20 Hz frequency range of the ECG signal.}
          \label{fig:cleanedsignal}
        \end{figure}

                ECG signal is frequently contaminated by a variety of noise sources, such as 50/60 Hz interference from power lines, EMG signal from muscles, motion artefacts, and variations in electrode-skin contact. Therefore, a band-pass filter with a frequency range of 8 to 12 Hz was applied to remove noise and artifacts while maintaining the ECG signal's QRS complex properties ~\cite{elgendi2010frequency}. The signals before and after passing the band-pass filter are shown in Fig.~\ref{subfig: Subfig3a} and Fig.~\ref{subfig: Subfig3b}, respectively.
        
    \subsection{Feature extraction}
        In order to detect ECG segments in patients with SA, abnormal characteristics of the ECG signal are extracted and fed into a machine learning classifier. As low QRS voltage and a leftward shift of the electrical axis may be associated with certain ECG abnormalities of SA patients~\cite{bacharova2015effect}, features associated to the QRS complex of the ECG signal can be utilized to detect SA episodes efficiently. Previous studies have classified ECG signals with and without SA based on characteristics associated to R peaks, which has limitations as other peaks are neglected (P, Q, S, T). To the best of our knowledge,  this study is the first to recover features linked to S peaks and then then feed them into a CNN model in order to distinguish between SA and non-SA ECG. We use the following features for the classification purpose:
        \begin{itemize}
            \item Amplitude of R peaks
            \item Amplitude of S peaks
            \item RR interval (duration between two consecutive R peaks)
            \item SS interval (duration between consecutive S peaks)
        \end{itemize}
        
\begin{figure}[hbt!]
          \centering
          \subfloat[]{\includegraphics[scale=0.6]{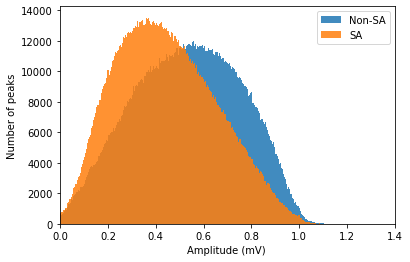}} \\
          \subfloat[]{\includegraphics[scale=0.6]{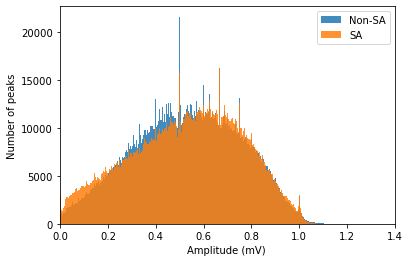}}
          \caption{\textit{(a)} The histogram of R peaks and  \textit{(b)} The histogram of RR intervals.}
          \label{fig:histogramR}
    \end{figure}

\begin{figure}[hbt!]
          \centering
          \subfloat[]{\includegraphics[scale=0.6]{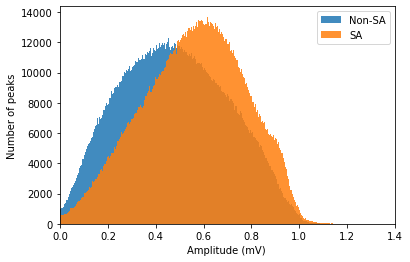}} \\
          \subfloat[]{\includegraphics[scale=0.6]{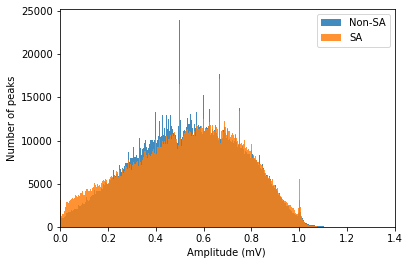}}
          \caption{\textit{(a)} The histogram of S peaks and \textit{(b)} The histogram of SS intervals.
          }
          \label{fig:histogramS}
    \end{figure}

        The distributions of R-peak-related features and S-peak-related features are depicted in Fig.~\ref{fig:histogramR} and Fig.~\ref{fig:histogramS}, respectively. 
        It seems that the RR and SS intervals of SA ECG are the same as those of Non-SA ECG. It is noteworthy that R peaks in SA ECG have significantly lower amplitude than those in Non-SA ECG. In contrast, S peaks in SA ECG have a higher amplitude than those in non-SA. These analyses suggest that these features can be used to distinguish between SA and Non-SA.
        
        To determine the positions of the R and S peaks, the following steps are taken:

        \begin{itemize}
            \item \textbf{\emph{R peaks detection:}} In order to determine the positions of the R peaks and calculate the RR intervals, we first used the Hamilton algorithm~\cite{hamilton2002open} to locate the R peaks. The amplitude of the R peaks are then extracted, and their positions are utilized to estimate the RR intervals. To remove redundant R peaks due to false detection, a local median filter proposed in~\cite{chen2014automatic} is applied. As the suspected irregular RR intervals can be caused by either false R peaks or missed R peaks, a lower bound and an upper bound are defined based on the physiological range of RR intervals in order to distinguish between these two types of uninterpretable data points. For abnormal RR intervals caused by false R peaks detection, the RR intervals in a sliding window are compared to the lower bound and rectified using either averaging or merging procedures. For irregular RR intervals caused by a missed R wave detection, the RR intervals in the current window are divided into several equal values or averaged with the neighboring window based on the specified criteria.
            
            \item \textbf{\emph{S peaks detection:}} A method for detecting S peaks based on the position of R peaks is proposed, in which, the resulting S peak is the first negative peak (in the case of a positive R peak). S peak detection algorithm is specified in algorithm~\ref{alg:s_peaks}.
            
            \begin{algorithm}[!hbt]
            \small{
            \begin{algorithmic}
                \Function{find\_S\_peaks}{data$\_$ecg, R$\_$peaks}
                    \State {num$\_$peaks $\gets$  R$\_$peaks}.shape[0]
                    \State {S$\_$peaks $\gets$ {list()}}
                    \State {$N$ $\gets$length{(num$\_$peaks)}}
                   \For{$index \gets 0$ to $N$}                    
                        \State {$i$ $\gets$ R$\_$peaks[index]}
                        \State {cnt $\gets$ $i$}
                        \If{cnt+1 $>=$  data$\_$ecg.shape[0]}
                            \State break
                        \EndIf

                        \While{data$\_$ecg[cnt] $>$ data$\_$ecg[cnt+1]}
                            \State {cnt $\gets$ cnt+1}
                                \If{cnt $>=$  data$\_$ecg.shape[0]}
                                    \State break
                                \EndIf
                        \EndWhile
                        \State {S$\_$peaks.append(cnt)}
                    \EndFor
                    \State \Return {S$\_$peaks}
                \EndFunction
            \end{algorithmic}
            \caption{Algorithm for determining the ECG S peaks}
            \label{alg:s_peaks}}
            \end{algorithm}
        \end{itemize}
        
\begin{figure}[hbt!]
        \scriptsize{
          \centering
          \subfloat[]{\includegraphics[scale=0.6]{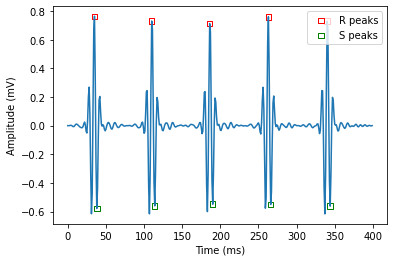} 
          \label{subfig: Subfig4a}} \\
          \subfloat[]{\includegraphics[scale=0.6]{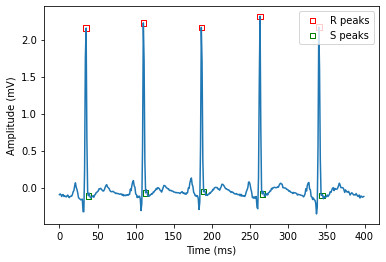}
          \label{subfig: Subfig4b}}
          \caption{Positions R and S peaks are detected on the original signal \textit{(a)} and on the filtered signal \textit{(b)}.}}
        \end{figure}

        Fig.~\ref{subfig: Subfig4a} and Fig.~\ref{subfig: Subfig4b} display the location of the R and S peaks found by the algorithm described above on the filtered signal and the corresponding raw signal, respectively.
        
            The effectiveness of the method for detecting R and S peaks is evaluated by manually labeling the positions of R and S peaks, followed by a comparison with the algorithm's return values. We randomly labeled 200 R peaks and 200 S peaks in both classes (with and without SA); the accuracy and F1-score of R and S peaks detection algorithms are reported in Table~\ref{table: evaluate}. All the statistics are performed on the \textit{PhysioNet Apnea–ECG dataset}. 
        
        \begin{table}[!hbt]
        \centering
        \small{
            \caption{R and S peaks detection results for filtered signal.}
            \begin{tabular}{|c|cc|c|c|c|c|c|c|}
                \hline
                 & \multicolumn{1}{c|}{\begin{tabular}[c]{@{}c@{}}\#beats \end{tabular}} & TP & TN & FP & FN & Acc & F1-score  \\ \hline 
                R peaks & \multicolumn{1}{c|}{\begin{tabular}[c]{@{}c@{}}200\end{tabular}} & 196 &0 & 0 & 4 & 98\% & 98.99\% \\ \hline 
                S peaks & \multicolumn{1}{c|}{\begin{tabular}[c]{@{}c@{}}200\end{tabular}} & 196 & 0 & 0 & 4 & 98\% & 98.99\% \\ \hline 
            \end{tabular}
            \label{table: evaluate}}
        \end{table}
        
        Due to the disparity in units (amplitude in millivolts and time interval in seconds), feature values are normalized to the normal distribution before feeding them into the neural network. Based on the findings on study~\cite{wang2019sleep}, cubic interpolation yielding 900 points of each feature every 5-minute segment is utilized as an effective data augmentation technique.
        
    
    \subsection{Performance metrics}
        In this study, to evaluate the performance of the machine learning classifier, we adopt the accuracy, sensitivity, specificity and F1-score as evaluation metrics.
        
        \begin{itemize}
            \item Sensitivity: Sensitivity of a class is defined as the ratio of correctly classified samples to total number of samples actually belonging to that class.
                \begin{equation}
                    \text{\emph{Sensitivity}} = \frac{TP}{TP+FN}
                \end{equation}
            \item  Specificity: Specificity is used to measure the proportion of negatives that are correctly identified. It is defined as the ratio of true negatives predicted to total number of samples which belong to negative class.\\
                \begin{equation}
                     \text{\emph{Specificity} } = \frac{TN}{TN+FP}
                \end{equation}
            \item Accuracy: It is defined as the ratio of number of correctly classified samples to that of total samples.\\
                \begin{equation}
                    \text{\emph{Accuracy}} = \frac{TP+TN}{TP+FN+TN+FP}
                \end{equation}
            \item F1-score: It  is the harmonic mean of the precision and recall. \\
                \begin{equation}
                    \text{\emph{F1-score}} = \frac{2TP}{2TP+FP+FN}
                \end{equation}
                
        \end{itemize}
        where \emph{TP}: True Positive, \emph{TN}: True Negative, \emph{FP}: False Positive, \emph{FN}: False Negative.\\
        
        
\section{Experiments}
\label{section:experiments}
    \subsection{Dataset}
    The PhysioNet Apnea--ECG dataset used to study SA was made available by Philipps University \cite{penzel2000apnea}. The dataset consists of 70 single-lead ECG records (35 recordings from the public set and 35 from the withheld set). Each recording ranges in length from just under 7 hours to over 10 hours and consists of a digitized ECG signal and a set of apnea annotations, which is derived by human experts on the basis of simultaneously recorded respiration and related signals. The ECG signal is separated into 1-minute segments and labeled; a signal segment is identified as having SA if an apnea event occurs during that minute. Table~\ref{table:dataset} provides details on the number of 1-minute ECG signal segments identified with and without SA in the training and test sets. 
    
    \begin{table}[!hbt]
    \centering

    \small{
        \caption{Number of 1-minute ECG signal segments labeled with and without SA on training and test sets.}
        \begin{tabular}{|c|c|c|}
        \hline
               & Training set      & Test set \\ \hline
        SA     &  6,473 (38.74\%)      & 6,490 (38.30\%)   \\ \hline
        Non SA &  10,236 (61.26\%)     & 10,455 (61.79\%) \\ \hline
        \textbf{Total}  &  \textbf{16,709}     & \textbf{16,945}      \\ \hline
        \end{tabular}
        \label{table:dataset}}
        \end{table}
    
    Previous research has demonstrated that adjacent segments provide valuable information for SA detection~\cite{yadollahi2009acoustic, wang2019sleep}. We adopt the sampling method in~\cite{wang2019sleep}, in which, each 1-minute signal segment and the surrounding 2-minute signal are used to form a 5-minute signal segment, which is then used for preprocessing and classification.
    
    \subsection{Implementation details and training methodology}
    After being preprocessed, the extracted features will be utilized to train the 1D CNN model. The NVIDIA GeForce RTX 3080 Ti GPU, 31 GB of RAM, and Intel Core i9-10900X processor operating at 3.70 GHz are used in all experiments. With a mini-batch size of 256, we train a specific model for 100 epochs, evaluating each model after every epoch. The optimal model for each training procedure will be the checkpoint with the highest F1-score. To assess the impact of employing features associated with S peaks, we set up two experiments as follows:
        \begin{enumerate}
            \item Only R peak amplitude and RR intervals, which are features related to R peaks, are used.
            \item Use a combination of features related to R and S peaks, including amplitudes and intervals of R and S peaks.
        \end{enumerate}
        
        The ECG signal classifier is constructed using the SE-ResNext~50~\cite{hu2018squeeze} model as its backbone. The SE-ResNext~50~\cite{he2016deep} model is a variant of ResNet~50 with the replacement of the identity connection with a Squeeze-and-Excitation block. It enables feature recalibration, allowing the network to learn how to use global information to selectively emphasize informative characteristics and suppress less helpful ones. This model has demonstrated its effectiveness with 1D data and ECG signals in particular~\cite{wang2020automatic}.
        
\section{Results and Discussion}
\label{section:results}
    
    \subsection{Model performance} As previously noted, in order to test the efficacy of using S peak-related features, we conducted experiments with and without these features. Table~\ref{tab:2_experiments} shows the classification results of the model employing solely R-related features versus the model employing all types of features. Our approach (RR intervals + R Amplitude) reports an accuracy of 90.28\%, a specificity of 90.44\%, and a sensitivity of 90.00\%. In terms of F1-score, we report an F1-score of 86.85 (for SA class) and 92.30 for non SA class. According to the results, using features associated to the S peak enhances the F1-score by up to 1.5\% in the SA class and 0.54\% in the Non-SA class. That finding indicates that the characteristics associated with the S peak may represent an alteration in the ECG signal of individuals with SA and may contribute to the capacity to classify SA and Non-SA.\\
    
    \subsection{Comparison to state-of-the-art} Table~\ref{tab:comparison} show the comparison between the proposed method and current state-of-the-art approaches to SA detection. We show that our approach surpassed almost all of the competitors, with the largest disparity in accuracy is up to 9.1\%. These results indicate the robustness of the propose method. 
    
    \begin{table}[!hbt]
    \scriptsize{
        \caption{Experimental results using different features.}
            \begin{tabular}{|ccccccc|}
            \hline
            \multicolumn{7}{|c|}{Results}\\ \hline
            \multicolumn{1}{|c|}{\begin{tabular}[c]{@{}c@{}}Feature\\ combination\end{tabular}}& \multicolumn{1}{c|}{Classifier}& \multicolumn{1}{c|}{Acc (\%)} & \multicolumn{1}{c|}{Spe (\%)} & \multicolumn{1}{c|}{Sen (\%)}
            & \multicolumn{2}{c|}{F1-score (\%)} \\ \cline{6-7} 
            \multicolumn{1}{|c|}{}  & \multicolumn{1}{c|}{}  & \multicolumn{1}{c|}{} & \multicolumn{1}{c|}{}& \multicolumn{1}{c|}{} & \multicolumn{1}{c|}{SA}      & \begin{tabular}[c]{@{}c@{}}Non \\ SA\end{tabular} \\ \hline
            \multicolumn{1}{|c|}{\begin{tabular}[c]{@{}c@{}}RR intervals\\ R Amplitude\end{tabular}}                              & \multicolumn{1}{c|}{\begin{tabular}[c]{@{}c@{}}SE\\ ResNext\\ 50\end{tabular}} & \multicolumn{1}{c|}{90.28}              & \multicolumn{1}{c|}{90.44}              & \multicolumn{1}{c|}{90.00}              & \multicolumn{1}{c|}{86.85} & 92.30  \\ \hline
            \multicolumn{1}{|c|}{\begin{tabular}[c]{@{}c@{}}RR intervals\\ R Amplitude\\ SS intervals\\ S Amplitude\end{tabular}} & \multicolumn{1}{c|}{\begin{tabular}[c]{@{}c@{}}SE\\ ResNext\\ 50\end{tabular}} & \multicolumn{1}{c|}{91.13}& \multicolumn{1}{c|}{92.58}              & \multicolumn{1}{c|}{88.75}              & \multicolumn{1}{c|}{88.35} & 92.84                                        \\ \hline
            \end{tabular}
            \newline\newline
            \label{tab:2_experiments}}
    \end{table}

    \begin{table}[!hbt]
    \centering
    \caption{Comparison with state-of-the-art approaches.}
        \begin{tabular}{|ccccc|}
            \hline
            \multicolumn{5}{|c|}{Comparison}\\ \hline
                \multicolumn{1}{|c|}{Study}&\multicolumn{1}{c|}{Classifer}&\multicolumn{1}{c|}{Accuracy}& \multicolumn{1}{c|}{Sensitivity}& Specificity\\ \hline
                \multicolumn{1}{|c|}{\begin{tabular}[c]{@{}c@{}}Li\\et al.\cite{li2018method}\end{tabular}}& \multicolumn{1}{c|}
                {\begin{tabular}[c]{@{}c@{}}Decision \\  fusion \end{tabular}}
                & \multicolumn{1}{c|}{83.80\%}& \multicolumn{1}{c|}{88.90\%} & 88.40\%\\ \hline
                \multicolumn{1}{|c|}{\begin{tabular}[c]{@{}c@{}}André Pinho\\et. al.\cite{pinho2019towards} \end{tabular}}& \multicolumn{1}{c|}{ANN}& \multicolumn{1}{c|}{82.12\%}& \multicolumn{1}{c|}{88.41\%}& 89.10\%\\ \hline
                \multicolumn{1}{|c|}{\begin{tabular}[c]{@{}c@{}}Mahsa Bahrami\\et. al.\cite{wang2019sleep}\end{tabular}}&\multicolumn{1}{c|}{LeNet-5}& \multicolumn{1}{c|}{87.6\%}& \multicolumn{1}{c|}{83.10\%}& 90.30\%\\ \hline
                
                \multicolumn{1}{|c|}{\begin{tabular}[c]{@{}c@{}}Shen \\ et. al. \cite{shen2021multiscale}  \end{tabular}}& \multicolumn{1}{c|}
                {\begin{tabular}[c]{@{}c@{}}1DCNN \\  WLTD \end{tabular}}
                & \multicolumn{1}{c|}{89.40\%}& \multicolumn{1}{c|}{89.80\%}& 89.10\% \\ \hline
                
                \multicolumn{1}{|c|}{\begin{tabular}[c]{@{}c@{}}Mahsa Bahrami \\ et. al. \cite{bahrami2022sleep}  \end{tabular}} & \multicolumn{1}{c|}
                {\begin{tabular}[c]{@{}c@{}}ZFNet \\  GRU \end{tabular}}
                & \multicolumn{1}{c|}{88.13\%}& \multicolumn{1}{c|}{84.26\%}& \textbf{92.27\%}\\ \hline
                
                \multicolumn{1}{|c|}{\begin{tabular}[c]{@{}c@{}}Singh \\ et. al. \cite{singh2019novel}  \end{tabular}}& \multicolumn{1}{c|}{Scalogram}& \multicolumn{1}{c|}{86.22\%}& \multicolumn{1}{c|}{90.00\%}& 83.82\%\\ \hline
                \multicolumn{1}{|c|}{\begin{tabular}[c]{@{}c@{}}Kaicheng Feng \\  et. al. \cite{feng2020sleep}  \end{tabular}}& \multicolumn{1}{c|}{FSSAE}& \multicolumn{1}{c|}{85.10\%}& \multicolumn{1}{c|}{86.20\%}& 84.40\% \\ \hline
                
                
                \multicolumn{1}{|c|}{\textbf{Ours [a]}}& \multicolumn{1}{c|}{\begin{tabular}[c]{@{}c@{}}SE-ResNext \\  50 \end{tabular}} & \multicolumn{1}{c|}{90.28\% }& \multicolumn{1}{c|}{90.44\%} & 90.00\% \\ \hline
                \multicolumn{1}{|c|}{\textbf{Ours [b]}}& \multicolumn{1}{c|}{\begin{tabular}[c]{@{}c@{}}SE-ResNext \\  50 \end{tabular}} & \multicolumn{1}{c|}{\textbf{91.13\%}} & \multicolumn{1}{c|}{\textbf{92.58\%}} & 88.75\% \\ \hline
        \end{tabular}
        \newline
        \newline
    {\raggedright a: The model solely uses features relating to R peaks (RR intervals and R amplitude).\\
    b: The model uses additional S-peak-related features (RR intervals, R amplitude, SS intervals and S amplitude).
    \par}
    \label{tab:comparison}   
    \end{table}

\section{Conclusions}
\label{section:conclusion}

In this study, we introduce a novel method for detecting sleep apenea based on SS intervals and S amplitudes. Our experimental results demonstrate that the proposed technique is effective for SA detection, and its performance outperforms state-of-the-art works. Although showing a high-level of performance, our approach has several disadvantages. For example, the lack of access to SA patients' ECG data and the large volume of data are the primary limitations of this study. Although proposed method has been tested with a number of experimental setups, it should be evaluated on a larger dataset. 

In the future, we expect to integrate a SA detection module into a mobile application, which will alert patients suffering from SA to wake them up. Additionally, we also consider muti-model learning approaches \cite{moro2019multimodal}, e.g. integrating demographic data and clinical data \cite{le2022enhancing} to boost model performance. An example can be the image and encoded text (in the form of a graph neural network or a knowledge graph) fusion for better representation learning \cite{tan2020multimodal,huang2019image,nguyen2022novel,nguyen2022image}. Not only that, we will use the explainable method to indicate which segments are at risk of disease and improve the accuracy of the model \cite{lekhiem}. Specifically, we will highlight which segments of the ECG signal are suspect in the ECG signal obtained using an eXplainable. As a result, this application may save the patient's life by restoring normal breathing and consciousness without complex setups.


\section{Acknowledgement }
    The authors would like to thank the VinUni-Illinois Smart Health Center and the source of funding : VinUni Seed Grant 2020. For more information, please visit the website at \url{https://vinuni.edu.vn/vinuni-research/seed-funding-program/}. Beside, this work was also funded by Vingroup Joint Stock Company (Vingroup JSC), Vingroup, and supported by Vingroup Innovation Foundation (VINIF) under project code VINIF.2021.DA00128.

\bibliography{sleepapena_ref.bib}
\end{document}